\documentclass[sigconf]{acmart}

\usepackage{booktabs} 

\newcommand{\Ni}{({\em i})~}
\newcommand{\Nii}{({\em ii})~}
\newcommand{\Niii}{({\em iii})~}
\newcommand{\Niv}{({\em iv})~}

\acmDOI{http://dx.doi.org/10.1145/3077136.3080757}

\acmISBN{978-1-4503-5022-8/17/08}

\acmConference{SIGIR '17}{}{August 07-11, 2017, Shinjuku, Tokyo, Japan}
\acmPrice{15.00}
\acmDOI{10.1145/3077136.3080757}
\acmYear{2017}
\copyrightyear{2017}

\acmPrice{15.00}

\begin{document}
\title{Large-Scale Goodness Polarity Lexicons\\
for Community Question Answering}
\renewcommand{\shorttitle}{Large-Scale Goodness Polarity Lexicons for cQA}

\author{Todor Mihaylov}
\affiliation{%
  \institution{Heidelberg University\\AIPHES Research Training Group}
  \streetaddress{Im Neuenheimer Feld 325}
  \city{Heidelberg} 
  \state{Germany} 
  \postcode{69120}
}
\email{mihaylov@cl.uni-heidelberg.de}

\author{Daniel Balchev,\\ Yasen Kiprov, Ivan Koychev}
\affiliation{%
  \institution{Sofia University}
  \streetaddress{5 James Bourchier blvd.}
  \city{Sofia} 
  \state{Bulgaria} 
  \postcode{1164}
}
\email{koychev@fmi.uni-sofia.bg}

\author{Preslav Nakov}
\affiliation{%
  \institution{Qatar Computing Research Institute, HBKU}
  \streetaddress{HBKU Research Complex}
  \city{Doha} 
  \state{Qatar} 
  \postcode{P.O. Box 5825}
}
\email{pnakov@hbku.edu.qa}

\renewcommand{\shortauthors}{Mihaylov et al.}

\begin{abstract}
We transfer a key idea from the field of sentiment analysis to a new domain: community question answering (cQA). The cQA task we are interested in is the following: given a question and a thread of comments, we want to re-rank the comments, so that the ones that are good answers to the question would be ranked higher than the bad ones. We notice that good vs. bad comments use specific vocabulary and that one can often predict the goodness/badness of a comment even ignoring the question, based on the comment contents only. This leads us to the idea to build a good/bad polarity lexicon as an analogy to the positive/negative sentiment polarity lexicons, commonly used in sentiment analysis.
In particular, we use pointwise mutual information in order to build large-scale goodness polarity lexicons in a semi-supervised manner starting with a small number of initial seeds.
The evaluation results show an improvement of 0.7 MAP points absolute over a very strong baseline, and state-of-the art performance on SemEval-2016 Task 3.
\end{abstract}

%
%

\ccsdesc[500]{Computing methodologies~Natural language processing}
\ccsdesc[500]{Information systems~Question answering}
\ccsdesc[100]{Information systems~Retrieval models and ranking}
\ccsdesc[100]{Information systems~Learning to rank}

\keywords{Community Question Answering, Goodness polarity lexicons, Sentiment Analysis.}

\maketitle

\section{Introduction}


Since the very early days of the field of sentiment analysis, 
researchers have realized that this task was quite different from other natural language processing (NLP) tasks such as document classification \cite{Sebastiani:2002:MLA:survey}, e.g., into categories such as business, sport and politics, and that it crucially needed external knowledge in the form of special sentiment polarity lexicons, which could tell the out-of-context sentiment of some words. See for example the surveys by \citet{PangL08} and \citet{Liu12} for more detail about research in sentiment analysis.


Intially, such sentiment polarity lexicons were manually crafted, and were of small to moderate size, e.g.,
LIWC \cite{pennebaker01},
General Inquirer \cite{inquirer1966computer},
Bing Liu's lexicon \cite{Hu04},
and MPQA \cite{Wilson05},
all have 2,000-8,000 words.
Early efforts in building them automatically also yielded lexicons of moderate sizes \cite{esuli:lrec2006,swn}.

However, recent results have shown that automatically extracted large-scale lexicons (e.g., with a million entries) offer important performance advantages, as confirmed at shared tasks on Sentiment Analysis on Twitter at SemEval 2013-2017 \cite{Semeval2013,rosenthal-EtAl:2014:SemEval,Rosenthal-EtAl:2015:SemEval,nakov-EtAl:2016:SemEval,rosenthal-farra-nakov:2017:SemEval,Nakov:2016rm,nakov:2016:WASSA2016}, where over 40 teams participated four years in a row. Similar observations were made in the Aspect-Based Sentiment Analysis task at SemEval 2014-2016 \cite{Semeval2014task4,pontiki-EtAl:2015:SemEval,pontiki-EtAl:2016:SemEval}. In both tasks, the winning systems benefited from using massive sentiment polarity lexicons \cite{MohammadKZ2013,Zhu_SemEval2014}.
These large-scale automatic lexicons are typically built using bootstrapping,
starting with a small set of seeds 
of, e.g., 50-60 words,
and sometimes even just from emoticons 
\cite{MohammadKZ2013}; recent work has argued for larger, domain-specific seeds \cite{jovanoski-pachovski-nakov:2016:COLING}

Here we transfer the idea from sentiment analysis to a new domain: community question answering (cQA). The cQA task we are interested in is this \cite{nakov-EtAl:2015:SemEval,nakov-EtAl:2016:SemEval:task3,nakov-EtAl:2017:SemEval}: given a question and a thread of comments, we want to rank the comments, so that the ones that are good answers to the question would be ranked higher than the bad ones. We notice that good vs. bad comments have specific vocabulary and that one can often predict goodness/badness even ignoring the question. This leads us to the idea to build a goodness polarity lexicon as an analogy to the sentiment polarity lexicons.

In particular, we use pointwise mutual information (PMI) to build large-scale lexicons in a semi-supervised manner starting with a small number of seeds.
The evaluation results on SemEval-2016 Task 3 \cite{nakov-EtAl:2016:SemEval:task3} show that using these lexicons yields state-of-the art performance, and an improvement of 0.7 MAP points absolute over a very strong baseline.



\section{PMI and Strength of Association}
\label{sec:PMI}

Pointwise mutual information (PMI) comes from the theory of information: given two random variables $A$ and $B$, the mutual information of $A$ and $B$ is the ``amount of information'' (in units such as bits) obtained about the random variable $A$, through the random variable $B$ \cite{Church:1990:WAN:89086.89095}. 



PMI is central to a popular approach for bootstrapping sentiment lexicons proposed by \citet{turney2002thumbs}. It starts with a small set of seed positive (e.g., \emph{excellent}) and negative words (e.g., \emph{bad}), and then uses these words to induce sentiment polarity orientation for new words in a large unannotated set of texts.
The idea is that words that co-occur in the same text with positive seed words are likely to be positive, while those that tend to co-occur with negative words are likely to be negative. To quantify this intuition, Turney defines the notion of \emph{semantic orientation} (\emph{SO}) for a term $w$ as follows:
\begin{equation}
SO(w) = pmi(w,pos) - pmi(w,neg)
\end{equation}

\noindent where $pos$ and $neg$ stand for any positive and negative seed word, respectively.

The idea was later used by other researchers, e.g., \citet{MohammadKZ2013} built several lexicons based on PMI between words and 
seed emotional hashtags, i.e., \#happy, \#sad, \#angry, etc. or positive and negative smileys.




\section{Goodness Polarity Lexicon}
\label{sec:method}

We use SO to build goodness polarity lexicons for \emph{Good}/\emph{Bad} comments in Community Question Answering forums. Instead of using positive and negative sentiment words as seeds, we start with comments that are manually annotated as \emph{Good} or \emph{Bad} (from SemEval-2016 Task 3 datasets \cite{nakov-EtAl:2016:SemEval:task3}). 

From these comments, we extract words that are strongly associated with \emph{Good} or \emph{Bad} comments. Finally, we use these words as seeds to extract even more such words, but this time using bootstrapping with unannotated data.

In sum, unlike in the work above, we do not do pure bootstrapping, but rather we have a semi-supervised approach, which works in two steps.

{\bf Step 1:} To come up with a list of words that signal \emph{Good}/\emph{Bad} comment, and it is not easy to come up with such words manually, we look for words that are strongly associated with the \emph{Good} vs. \emph{Bad} comments in the annotated training dataset (where comments are marked as \emph{Good} vs. \emph{Bad}), using SO. We then select the top 5\% of the words with the most extreme positive/negative values of SO; this corresponds to the most extreme \emph{Good}/\emph{Bad} comment words.

{\bf Step 2:} We apply the SO again, but this time using the seed words selected in Step 1, to build the final large-scale goodness polarity lexicon, as in the above-described work.

Compared to previous work in sentiment analysis lexicon induction, we do not start with a small set of seed words, but rather with a set of comments annotated as \emph{Good} vs. \emph{Bad}, from which we extract \emph{Good}/\emph{Bad} seed words (using SO). Once we have these seed words, we proceed as is done for sentiment analysis lexicon induction (again using SO).

\section{Experiments and Evaluation}

We build a system that uses variety of features and is competitive to the best systems in the SemEval-2016 Task 3 competition; we then augment it with features extracted from our PMI-based goodness polarity lexicon.
We train an SVM classifier, where we create a training instance for each question-answer pair. 
Finally, we rank the comments for a given question based on the SVM score.

\subsection{Data}

We used the data from SemEval-2016 Task 3, Subtask A \cite{nakov-EtAl:2016:SemEval:task3}. It includes 6,398 training questions with 40,288 comments, plus an unannotated dataset comprising 189,941 questions and 1,894,456 comments. We performed model selection on the development dataset: 244 questions and 2,440 answers. The test dataset from the task, which we used for evaluation, included 329 questions and 3,270 comments.


\subsection{Non-lexicon Features}

We used several semantic vector similarity and metadata features, which we describe below.


\textbf{Semantic Word Embeddings.}
We used semantic word embeddings \cite{mihaylov-nakov:2016:SemEval} trained using word2vec\footnote{\url{https://github.com/tbmihailov/semeval2016-task3-cqa}} on the training data plus the unannotated Qatar Living data that was provided by the task organizers.
We also used embeddings pre-trained on GoogleNews \cite{word2vec}. 
For each piece of text such as comment text, question body and question subject, we constructed the centroid vector from the vectors of all words in that text (after excluding the stopwords).



\textbf{Semantic Vector Similarities.} We used various cosine similarity features calculated using the centroid word vectors on the question body, on the question subject and on the comment text, as well as on parts thereof:

\emph{Question to Answer similarity.} We assume that a relevant answer should have a centroid vector that is close to that for the question. We used the question body to comment text, and question subject to comment text vector similarities.

\emph{Maximized similarity.} We ranked each word in the answer text to the question body centroid vector according to their similarity and we took the average similarity of the top $N$ words. We took the top 1, 2, 3 and 5 word similarities as features. The assumption here is that if the average similarity for the top $N$ most similar words is high, then the answer might be relevant.

\emph{Aligned similarity.} For each word in the question body, we chose the most similar word from the comment text and we took the average of all best word pair similarities.

\emph{Part of speech (POS) based word vector similarities.} We performed part of speech tagging using the Stanford tagger \cite{Toutanova:2003:FPT:1073445.1073478}, and we took similarities between centroid vectors of words with a specific tag from the comment text and the centroid vector of the words with a specific tag from the question body text. The assumption is that some parts of speech between the question and the comment might be closer than other parts of speech.

\textbf{Word cluster similarity.} We first clustered the word vectors from the word2vec vocabulary into 1,000 clusters using K-Means clustering, which yielded clusters with about 200 words per cluster on average. 
We then calculated the cluster similarity between the question body's word clusters and the answer's text word clusters. For all experiments, we used clusters obtained from the word2vec model trained on the QatarLiving data with vector size 100, window size 10, minimum word frequency 5, and skip-gram context size 1.

\textbf{LDA topic similarity.} We performed topic clustering using Latent Dirichlet Allocation (LDA) 
of the questions and of the comments. We built topic models with 100 topics. For each word in the question body and for the comment text, we built a bag-of-topics with corresponding distribution, and we calculated similarity. The assumption here is that if the question and the comment share similar topics, they should be more likely to be relevant with respect to each other.

\textbf{Metadata.} In addition to the semantic features described above, we also used some features based on metadata:

\emph{Answer contains a question mark.} If the comment contains a question mark, it may be another question, which might indicate a bad answer.

\emph{Answer length.} The assumption here is that longer answers could bring more useful detail.

\emph{Question length.} If the question is longer, it may be more clear, which may help users give a more relevant answer.

\emph{Question to comment length.} If the question is long, but the answer is short, it is typically less relevant.

\emph{The answer's author is the same as the question's author.} It is generally unlikely that the user who asked the question would later on provide a good answer to his/her own question; rather, if s/he takes part in the discussion, it is typically for other reasons, e.g., to give additional detail, to thanks another user, or to ask additional questions \cite{nicosia-EtAl:2015:SemEval}.

\emph{Answer rank in the thread.} 
The idea is that discussion in the forum tends to diverge from the original question over time.

\emph{Question category.} We took the category of the question as a sparse binary feature vector.
The assumption here is that the question-comment relevance might depend on the category of the question.

\subsection{Goodness Polarity Lexicon Features}

We bootstrapped a goodness polarity lexicon using PMI as described above. This yielded a lexicon\footnote{Our goodness polarity lexicon is freely-available in the following URL:\\
\url{https://github.com/dbalchev/models/}} of 41,663 words, including 11,932 \emph{Bad} and 29,731 \emph{Good} words, with corresponding weigths, which describe the stregth of association of a word with \emph{Good} and \emph{Bad} comments: positive and negative weights, respectively. 
The \emph{Good} and the \emph{Bad} words with most extreme weights are shown in Table~\ref{tab:word-pmi-bootstrapped}. We can see that the \emph{Good} words mostly refer to locations, which is expected, e.g., for questions asking where something is located. In contrast, the \emph{Bad} words are mostly typos, names, numbers, and words in a foreign language.

Based on the goodness polarity lexicon, we extracted the following features for a target comment:
	\Ni number of \emph{Good} and \emph{Bad} words;
	\Nii number of \emph{Good} (and \emph{Bad}) / number of \emph{Good+Bad} words;
	\Niii sum of the scores of the \emph{Good}, sum of \emph{Bad} words, and sum of teh scores for \emph{Good+Bad} words;
	\Niv the highest score for a \emph{Good} word, and the lowest score for a \emph{Bad} word in the answer.

\begin{table}
\centering
\small
\begin{tabular}{|cc|cc|}
\hline
{\bf word} & {\bf SO} & {\bf word} & {\bf SO}\\
\hline
hayat & 6.917 & 13228 & -5.522\\
flyover & 6.195 & tting & -4.999\\
codaphone & 6.148 & illusions & -4.976\\
najada & 6.145 & bach & -4.849\\
rizvi & 6.107 & messiah & -4.566\\
emadi & 5.890 & dnm & -4.417\\
passportdept & 5.868 & daf & -4.356\\
omran & 5.728 & 2905 & -4.328\\
condenser & 5.698 & xppg & -4.313\\
bookstore & 5.688 & 29658 & -4.306\\
azzu & 5.634 & scorn & -4.219\\
5552827 & 5.634 & skamu & -4.053\\
overalling & 5.621 & rizk & -4.041\\
muncipilty & 5.538 & fiddledeedee & -3.954\\
\hline
\end{tabular}
\caption{The words with the biggest and the smallest SO scores
from our goodness polarity lexicon.}
\label{tab:word-pmi-bootstrapped}
\end{table}


\begin{table}[t]
\centering
\small
\begin{tabular}{|l|c|c|c|}
\hline
System & {\bf MAP} & AvgRec & MRR\\\hline
SemEval 1st & 79.19 & 88.82 & 86.42\\
\textbf{\em Our, with PMI lexicons} & \textbf{\em 78.75} & \textbf{\em 88.64} & \textbf{\em 86.69}\\
\textbf{\em Our, no PMI lexicons} & \textbf{\em 78.08} & \textbf{\em 88.37} & \textbf{\em 85.19}\\
SemEval 2nd & 77.66 & 88.05 & 84.93\\
SemEval 3rd & 77.58 & 88.14 & 85.21\\\
$\ldots$ & $\ldots$ & $\ldots$ & $\ldots$\\
Average & 73.54 & 84.61 & 81.54\\
$\ldots$ & $\ldots$ & $\ldots$ & $\ldots$\\
SemEval 12th (Worst) & 62.24 & 75.41 & 70.58\\\hline
Baseline$_{time}$   & 59.53 & 72.60 & 67.83 \\
Baseline$_{rand}$ & 52.80 & 66.52 & 58.71 \\\hline
\end{tabular}
\caption{Our results compared to those at SemEval, and to two baselines: chronological and random.}\label{tab:comparison}
\end{table}

\subsection{Results}

The evaluation results are shown in Table~\ref{tab:comparison}.
We can see that our system without goodness polarity lexicons would rank second on MAP and AvgRec, and third on MRR, at SemEval-2016 Task 3.
It outperforms a random baseline (Baseline$_{rand}$)
and a chronological baseline that assumes that early comments are 
better than later ones (Baseline$_{time}$) by large margins: by about 19 and 
25 MAP points absolute (and similarly for the other two measures).
It is also well above the worst and the average systems.
I.e., this is a very strong system, and thus it is not easy to improve over it.
Yet, adding the goodness lexicon features yields about 0.7 points absolute improvement in MAP;
the resulting system would have ranked second on MAP and AvgRec, and first on MRR.

%


\section{Conclusion and Future Work}

We have presented experiments in transferring an idea from sentiment analysis to a new domain: community question answering. In particular, 
we built
a goodness polarity lexicon 
that can help predict whether a comment is likely to be good or bad, regardless of the question asked. We have shown that using the lexicon yielded a sizeable improvement of 0.7 MAP points absolute over a very strong system, and near state-of-the art performance on SemEval-2016 Task 3.

In future work, we plan to extend the lexicon with $n$-grams.
We are further interested in trying other approaches for building polarity lexicons that go beyond PMI, e.g., using weights in SVM \cite{severyn2015automatic}; there was a special task on that topic at SemEval-2016 \cite{SemEval:lexicons}. We also plan to explore the impact of the quality of the words we use as seeds \cite{jovanoski-pachovski-nakov:2016:COLING}.


\vspace{-3pt}
\begin{acks}
The work is supported by the NSF of Bulgaria under Grant No.: DN 02/11/2016 - ITDGate.
\end{acks}

\bibliographystyle{ACM-Reference-Format}
\bibliography{sigproc} 

\end{document}